\pdfoutput=1

\documentclass[11pt,table,xcdraw]{article}

\usepackage{acl}

\usepackage{xcolor} 
\usepackage{times}
\usepackage{latexsym}
\usepackage{graphicx}
\usepackage{booktabs}
\usepackage{amsmath}
\usepackage{amsfonts}
\usepackage{setspace} 
\usepackage{url}
\usepackage[vlined]{algorithm2e}
\usepackage{xspace}

\usepackage[T1]{fontenc}

\usepackage[utf8]{inputenc}

\usepackage{microtype}
\usepackage{silence}  
\title{Can In-context Learners Learn a Reasoning Concept\\ from Demonstrations?}

\author{Michal Štefánik \and Marek Kadlčík \\
  Faculty of Informatics, Masaryk University, Czech Republic \\
  \texttt{\{stefanik.m,kadlcik\}@mail.muni.cz}
\vspace*{-2\baselineskip}}

\begin{document}
\maketitle
\begin{abstract}

Language models exhibit an emergent ability to learn a new task from a small number of input-output demonstrations.
However, recent work shows that in-context learners largely rely on their pre-trained knowledge, such as the sentiment of the labels, instead of learning new associations from the input.
We argue that the commonly-used few-shot evaluation using a \textit{random} selection of in-context demonstrations can not disentangle models' reliance on such biases, as most of the randomly-selected demonstrations do not present relations \textit{informative} for prediction beyond exposing the task's input-output distribution.

Therefore, to evaluate models' in-context learning ability independent of models' memory, we introduce a Concept-sharing few-shot learning method choosing the demonstrations that \textit{share} an underlying \textit{concept} with the predicted sample. We extract a set of such concepts from available human explanations and measure how much models can benefit from presenting these concepts in few-shot demonstrations.

We find that most of the recent in-context learners can not consistently benefit from the demonstrated concepts, irrespective of the model size. However, we note that \textsc{T0} models are more sensitive to exhibited concepts, benefiting from concept-sharing demonstrations in 7 out of 8 evaluation scenarios.

\end{abstract}

\section{Introduction}
\label{sec:intro}

\begin{figure}[tbh]
    \centering
    \hspace*{-2mm}
    \includegraphics[width=0.5\textwidth]{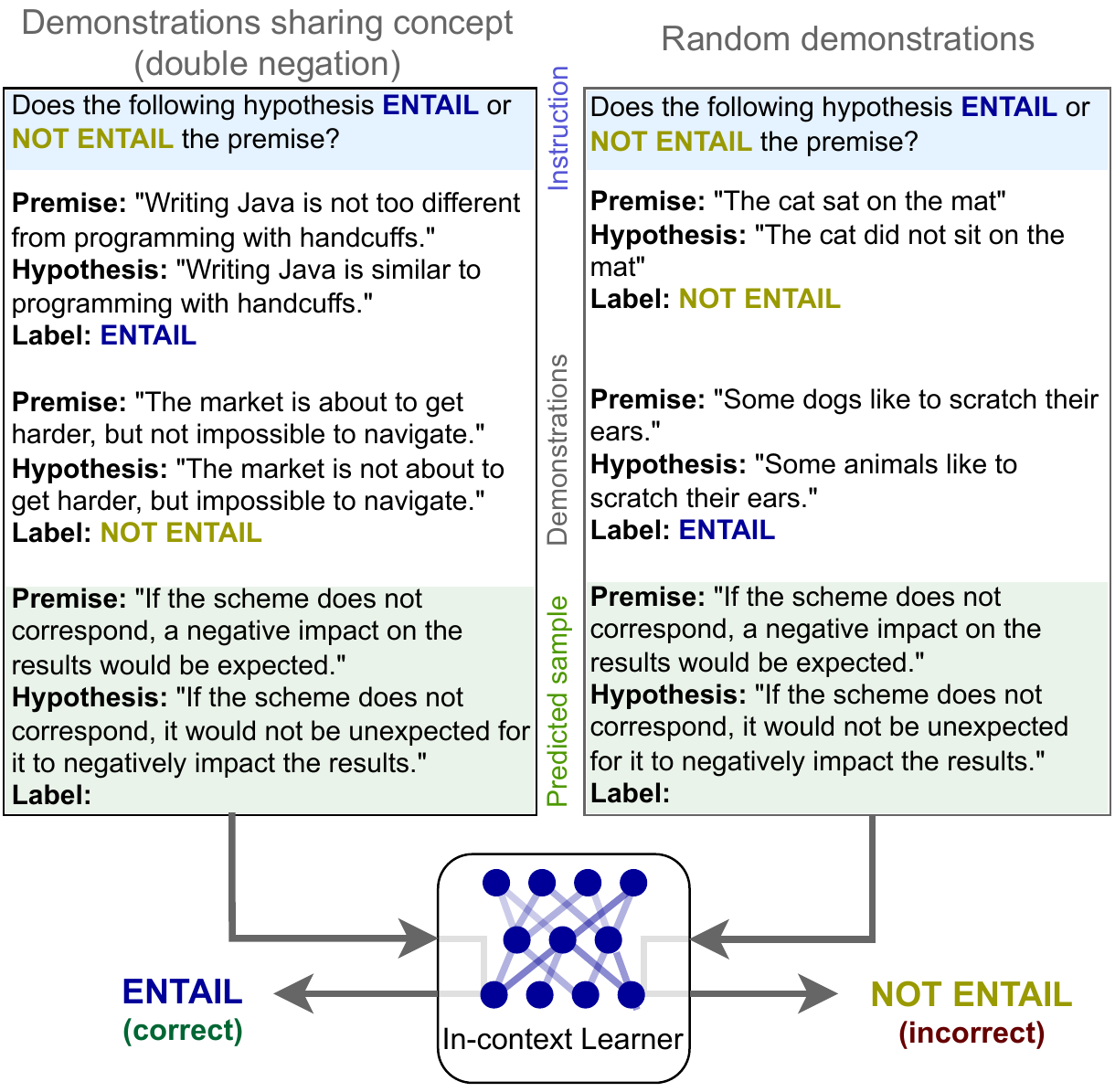}
    \caption{In this work, we assess in-context learners' ability to improve when presented with demonstrations using a reasoning concept applicable in the prediction (§\ref{sec:conceptual_fewshot}). We extract these concepts from human explanations (§\ref{sec:datasets}) and assess models' ability to learn to use these concepts, as reflected in improving their prediction quality.}
    \label{fig:abstract}
\end{figure}

In-context learning (ICL) is the alternative to the conventional training of Large Language Models (LLMs) for specific task(s), where models are expected to learn a new task solely from the input text. In few-shot in-context learning that we focus on, the input text contains a set of \textit{demonstrations}, i.e. the input-output examples of the task to be learned \cite{NEURIPS2020_brown_gpt3}.

An ability to learn unseen tasks from input sequence without updating the model has practical and theoretical implications, both of which are of great significance; Understanding free-form user requests allow applying LLMs in applications of restricted, or limited data availability without over-specialization \cite{Goodfellow2014AnEI}. In-context learning can provide a \textit{handle} of models' behaviour, enabling the model to avoid specific erroneous predictions. In theory, a training process resulting in an accurate new-task learner defines the sufficient conditions for the emergence of a specific level of generalization.

Recent LLMs trained on vast mixtures of tasks \citep{sanh2022multitask,wang-etal-2022-super,chung2022_flan} show a certain level of new-task ICL and gradually bring more attention and expectations in this direction. However, counter-intuitively to the overall evaluations, in-context learners (ICLs) also expose surprising behavioural artefacts; \citet{liu-etal-2022-makes} show ICLs' sensitivity to the ordering of in-context demonstrations. Similarly, \citet{lu-etal-2022-fantastically} find surprising sensitivity of ICLs to the specific wording of the prompts. \citet{what-makes-incontext-work} show that most of the model performance is persisted even when the contents of the demonstrations are randomly swapped. Contrary to the ability to learn from input, \citet{wei2023larger} propose to attribute this to the over-reliance of in-context learners on \textit{semantics} of the label tokens, especially in smaller models.

We hypothesize that the discrepancy between the expected and the perceived abilities of ICLs might be attributed to their limited evaluation, commonly performed with a \textit{random} set of task demonstrations.
However, for many open-ended tasks, such as question answering, or translation, randomly-chosen demonstrations rarely present a reasoning pattern which can help with the prediction of new input (Figure~\ref{fig:abstract}).
We argue that the evaluation with mostly non-informative contexts also can not reflect on the ability of \textit{learning}, as observed in humans\footnote{We restrain from discussing a concept of \textit{learning} in the psychological scope, but we note that Concept learning fits well into a definition of Associative learning \citep{plotnik2012}.},
as the gain of extrapolating associations presented in non-informative demonstrations can only bring little benefit to the practice.

We also note that in the absolute numbers, the random-demonstrations evaluation may favour some LLMs, such as ones with a capacity to remember a wider variety of input distributions from pre-training, that can be used for modulating their behaviour in ICL. However, note that such behaviour differs from learning new association(s) from the context and makes the model prone to adversaries.

Hence, in Section~\ref{sec:conceptual_fewshot}, we propose to evaluate models' in-context learning ability primed with the demonstrations that exhibit a reasoning \textit{analogical} to the one required for a robust prediction of the predicted sample (Figure~\ref{fig:abstract}). We measure how well can the recent few-shot learners \textit{utilize} identified concepts for more accurate predictions (§\ref{sec:evaluation}) and find large discrepancies among the models and concepts.

Our main contributions are following:
(i)~We introduce a task of Concept-sharing few-shot learning, disentangling models' ability to learn a new reasoning concept from other aspects of prediction quality. We show how such reasoning concepts can be extracted from human explanations.
(ii)~For a wide variety of recent in-context learners, we measure the ability to benefit from presented reasoning concepts. We show that while some models are better at learning concepts on average, this ability can not be attributed to the models' size or training strategy.

\paragraph{Problem Definition}
\label{sec:problem_def}

Given a dataset $\mathcal{D}: \{(x_1\rightarrow Y_1), .., (x_i\rightarrow Y_i)\} \in \mathcal{D}$ containing pairs of \textit{input} $x_j$ with associated \textit{label} $Y_j$, an \textit{in-context few-shot learner} $\Theta(x) \rightarrow y$ aims to predict a correct label $y_{k+1} = Y_{k+1}$ given a sequence of $k$ input-output \textit{demonstrations}, and the \textit{predicted input} $x_{k+1}$:
\begin{equation}
    \Theta([x_1\rightarrow Y_1, .., x_k\rightarrow Y_k], x_{k+1}) \rightarrow y_{k+1}
    \label{eq:few_shot}
\end{equation}
We expect \textit{in-context few-shot learner} $\Theta$ to model the relation of $x_{i}$ and $y_i$ by (i)~\textit{identifying} and (ii)~\textit{applying} the relations of input and output presented in demonstrations. Each such relation is modelled by one or more \textit{latent concepts} $\mathcal{C}$:
\begin{equation}
    \forall \, (x_i, Y_i) \in \mathcal{D}: \exists \, \mathcal{C}: \mathcal{C}(x_i, Y_i) = 1
    \label{eq:concept_def}
\end{equation}
We broadly define a \textit{concept} $\mathcal{C}$ as any function $\mathcal{C}(x, y) \rightarrow \{0, 1\}$, constraining a space of valid outputs $y$ to the ones where $\mathcal{C}(x, y) = 1$.
Thus, if $\Theta$ \textit{learns} a concept $\mathcal{C}$, it will never predict for $x$ such $y$ that $\mathcal{C}(x, y)\!=\!0$. In a composition $\{\mathcal{C}\}\!=\!\{\mathcal{C}_1,.., \mathcal{C}_j\}$, all $\mathcal{C}_i\!\in\!\{C\}$ must evaluate to~$1$.

Given that modelling of each $C$ valid for the task of $\mathcal{D}$ restrain a set of possible predictions of $\Theta$ \textit{exclusively} from incorrect predictions, extending a set of concepts learned in-context with complementary one(s) should \textit{never} decrease the performance of the model $\Theta$ on $\mathcal{D}$.

\section{Concept-sharing Few-shot Learning}
\label{sec:conceptual_fewshot}
We reformulate in-context few-shot learning \eqref{eq:few_shot} to a \textit{Concept-sharing few-shot learning}, evaluating the ability of a few-shot learner $\Theta$ to identify and apply a user-chosen reasoning concept $\mathcal{C}$ shown in demonstrations. First, we \textit{classify} evaluation samples such that the samples of the same category $X_i$ require the concept $\mathcal{C}_i$ to map $x$ to $Y$.
Subsequently, in concept-sharing few-shot learning, we let the learner to infer a prediction for input $x_{k+1}$ by presenting it with demonstrations $(x_j \rightarrow Y_j)_{1..k} \in X_i$, thus \textit{sharing} the reasoning concept $\mathcal{C}_i$ with the predicted input $x_{k+1}$:
\begin{equation}
    \begin{aligned}
        &\Theta([x_1\rightarrow Y_1, .., x_k\rightarrow Y_k], x_{k+1}) \\
        \text{where } &\forall (x_{1..k}, Y_{1..k})\in X^i \text{ and } x_{k+1} \in X^i
    \end{aligned}
    \label{eq:conceptual_few_shot}
\end{equation}
We note that $\Theta$ can rely on other features than $\mathcal{C}_i$, and such reliance is not easy to disentangle. Therefore, we propose to contextualize the results of Concept-sharing few-shot learning on a concept $C_i$ as a \textit{difference} to the performance obtained in a \textit{random} selection of demonstrations.

Additionally, to make the predictions based on two different sets of demonstrations mutually comparable without systematic bias (e.g. in samples' complexity), we perform both random and concept-sharing evaluations with the same predicted samples $x_{k+1}$, and only change the demonstrations.\footnote{The implementation of Concept-sharing few-shot learning is available on \url{https://github.com/MIR-MU/CoAT}.}


\paragraph{Informative Concepts Extraction}

Constructing a scaled evaluation with annotated reasoning concepts $\mathcal{C}$ is challenging since the annotations of such concepts among datasets are very rare.
However, we find the concepts inherently captured in human explanations of some datasets, where annotators are asked to collect answers to a question ``why is [input] assigned [output]?'' \citep{wiegreffe-marasovic-2021-review}.

The form of these explanations ranges from free-text explanations, including annotator-specific slang and stylistics, to semi-structured and structured explanations, cast to a pre-defined format, often consisting of a set of relations in a form ``[subject1] [relation] [subject2]'' that transitively maps [input] to [output] \citep{jansen-etal-2018-worldtree}.
We focus on extracting the concepts from the subset of the semi-structured and structured explanations where the format consistency ensures that analogical reasoning concepts will be shared among multiple samples.

\section{Evaluations}
\label{sec:evaluation}

This section overviews few-shot in-context learners that we evaluate for Concept-sharing few-shot learning and the datasets with explanations allowing us to extract shared reasoning concepts.

\subsection{Few-shot Learners}

\paragraph{\textsc{T0}} \citet{t0} introduce a set of in-context learning models fine-tuned from a T5 model \cite{t5} on a variety of 35 mostly QA tasks in zero-shot settings, aiming to perform well on a task of previously-unseen categories. T0 is trained for sequence-to-sequence generation over a large set of diverse tasks cast to a unified input-output format provided by task-specific templates of Promptsource \cite{bach2022promptsource}.

\paragraph{\textsc{Tk-instruct}} \citep{natural-instructions-2} is a set of models trained for comprehension of annotator-like instructions, consisting of a free-text task description and a set of input-output pairs, collected for over 1,600 tasks of \textsc{NaturalInstructions} collection~\citep{mishra-etal-2022-cross}.
In comparison to \textsc{T0}, \textsc{Tk-instruct} models can advance from a wider mix of tasks and also from fine-tuning in the few-shot learning format matching the evaluation.

\paragraph{\textsc{Flan}} \citep{https://doi.org/10.48550/arxiv.2210.11416} scales the approach of fine-tuning in a few-shot learning format to over 1,800 tasks of 146 categories including all resources of \textsc{T0} and \textsc{Tk-instruct}. Contrary to the former models, the training data mixture includes several datasets with chain-of-thought labels, where the model is trained to follow the annotated reasoning chain explicitly. We evaluate all publicly available \textsc{T5}-based \textsc{Flan} models.

\paragraph{\textsc{GPT3}} \citep{NEURIPS2020_brown_gpt3} is a well-known causal language model that has first shown that in-context few-shot learning ability can emerge solely from vast amounts of unsupervised training data and parametrization, without fine-tuning. Alternatively to other approaches, \textbf{\textsc{InstructGPT}} \cite{ouyang2022training} fine-tunes \textsc{GPT3} to follow human instructions using obtained user feedback. We evaluate both these models through OpenAI APIs\footnote{\url{https://beta.openai.com}}.



\subsection{Datasets}
\label{sec:datasets}

\begin{figure*}[tbh]
  \centering
    \hspace*{-8.1mm}
    \includegraphics[width=0.281\textwidth]{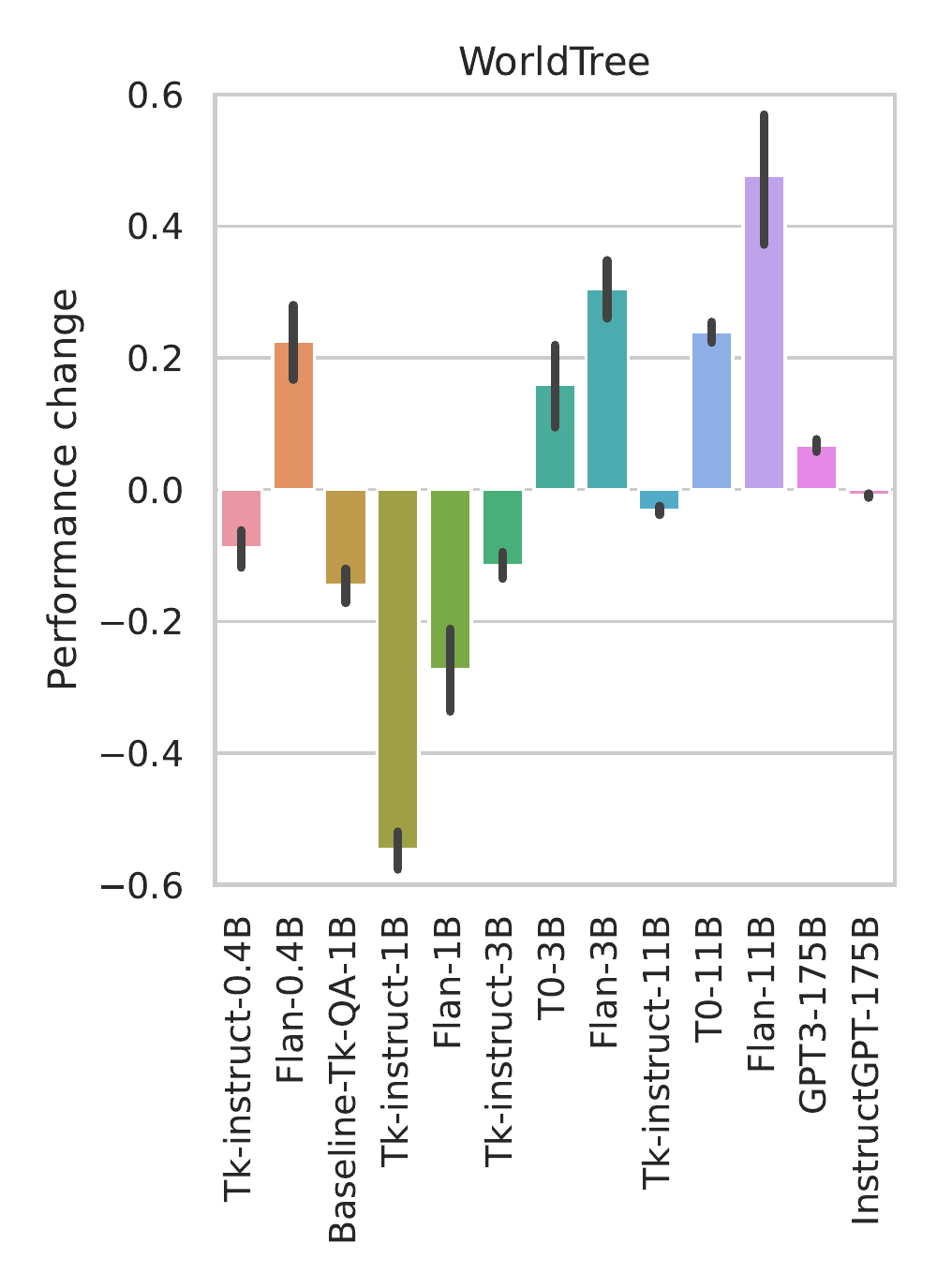} \hspace{-3mm}\includegraphics[width=0.281\textwidth]{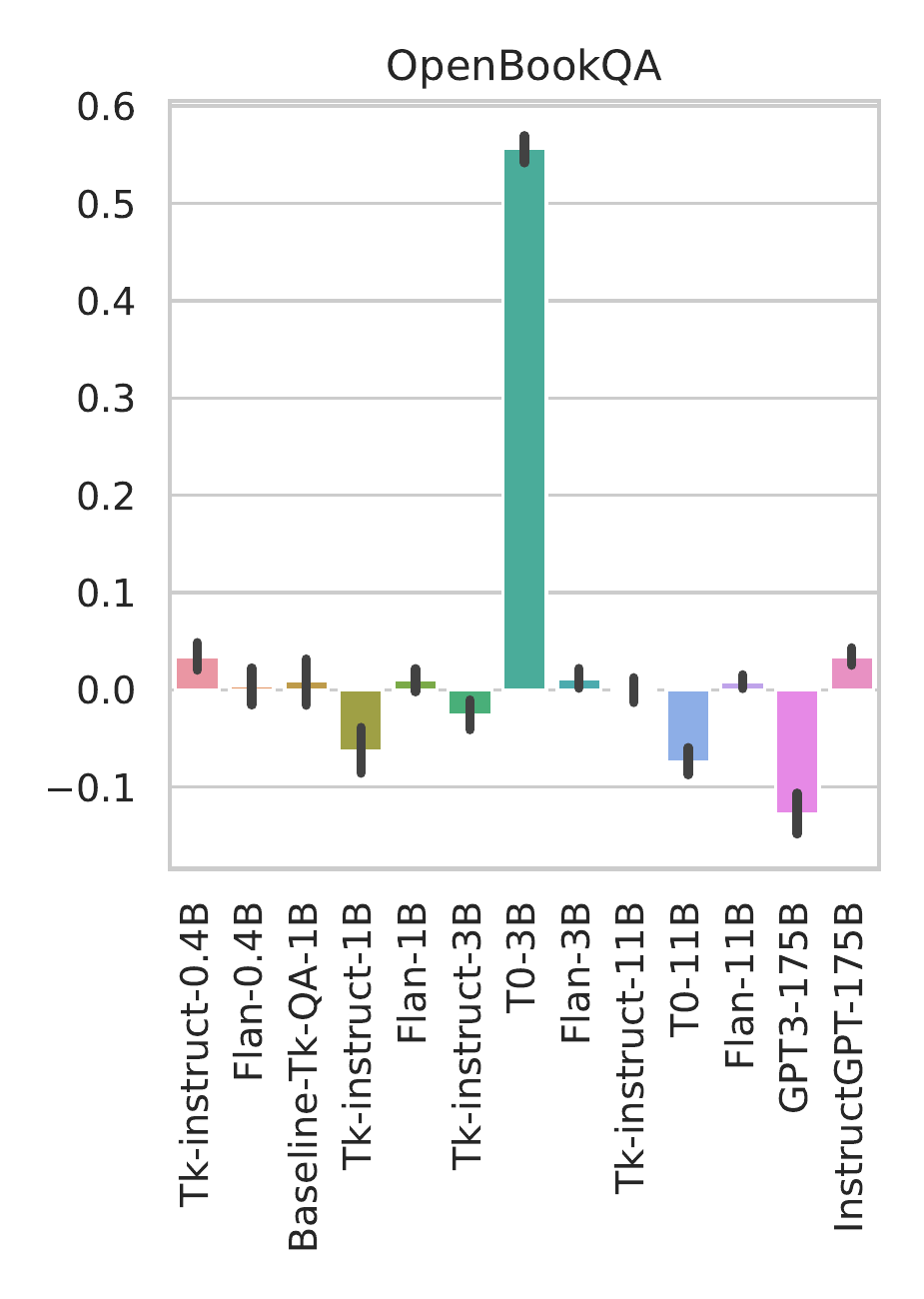} \hspace{-3mm}\includegraphics[width=0.241\textwidth]{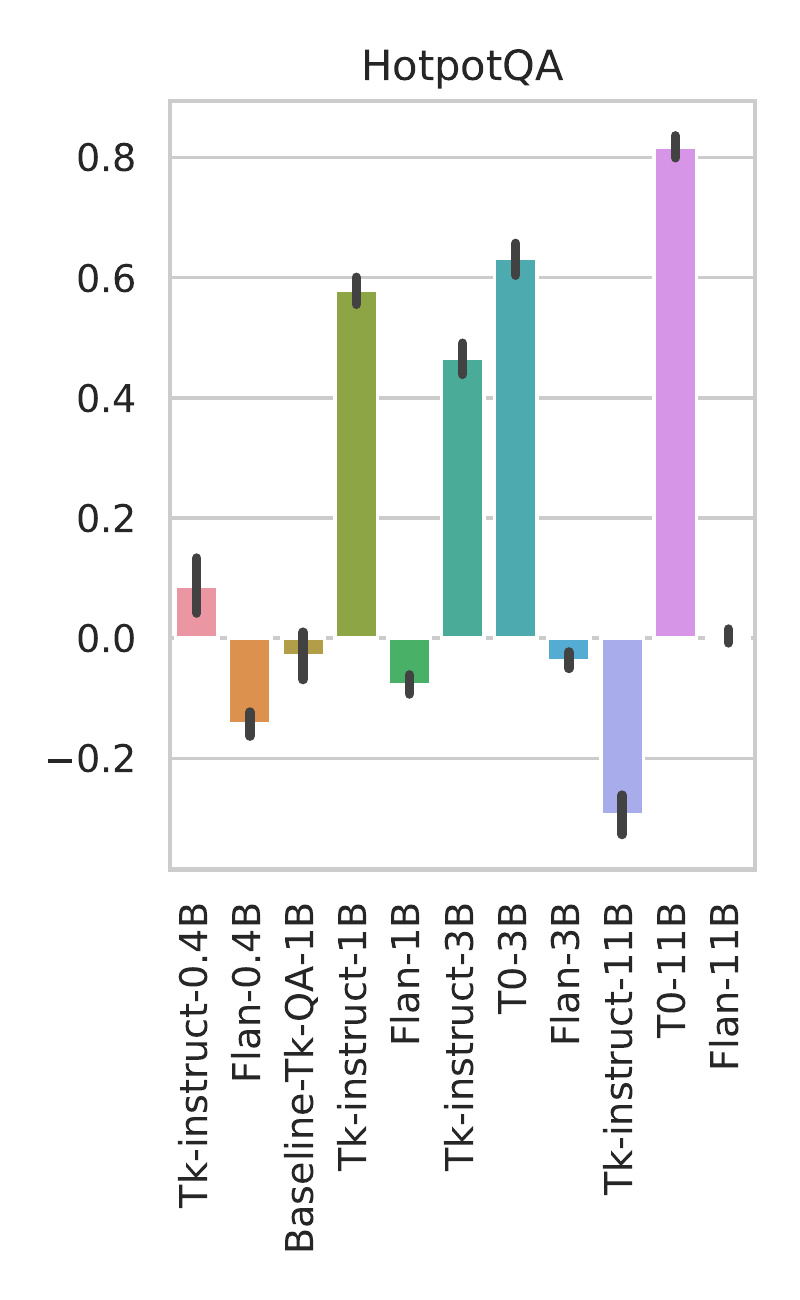}
    \hspace{-3mm}\includegraphics[width=0.281\textwidth]{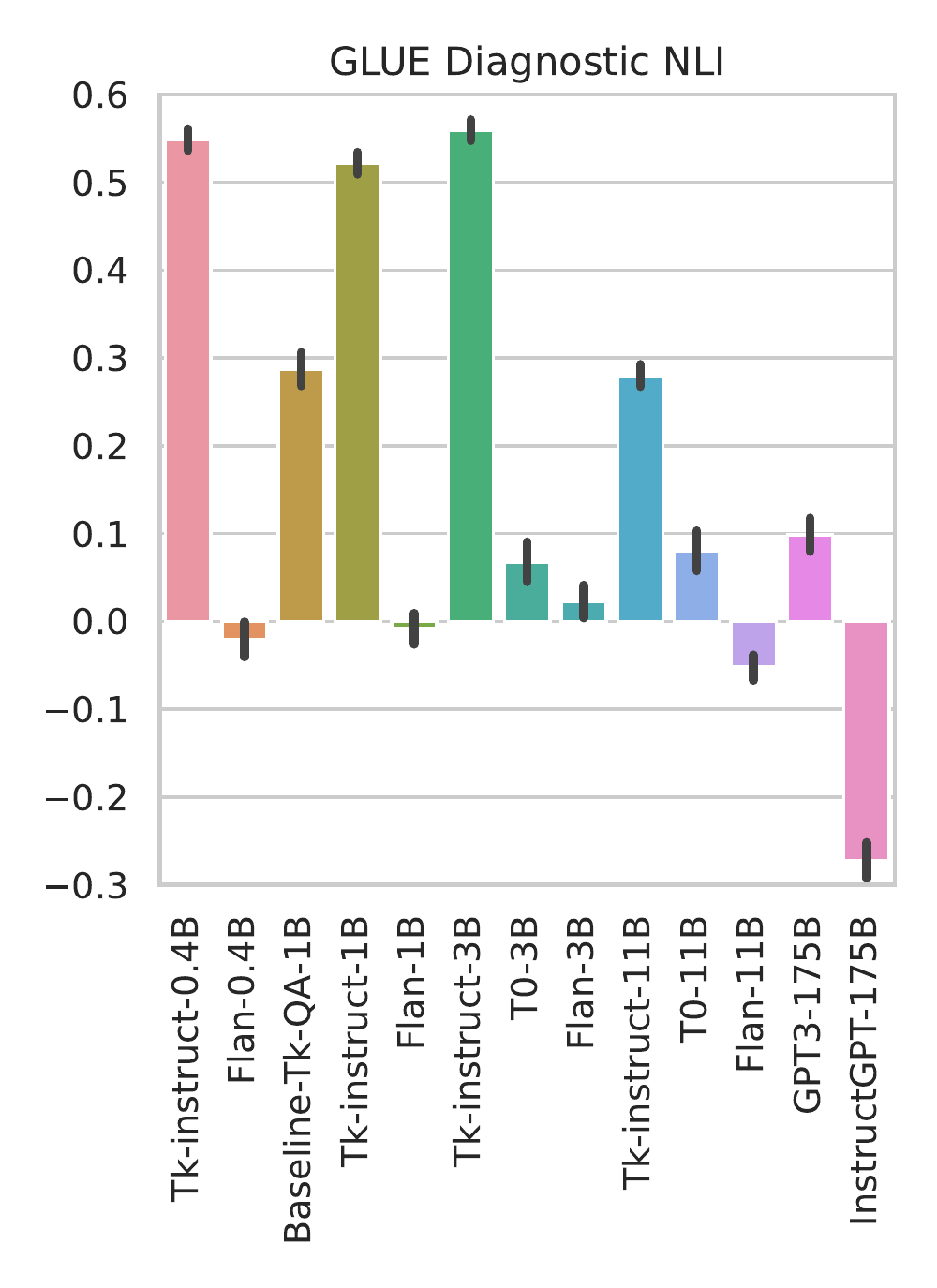}
  \hspace*{-5mm}
  \caption{\textbf{Conceptual few-shot evaluation:} Relative performance change of the assessed in-context learners between using \textit{random} demonstrations (k=3) and \textit{concept-sharing} demonstrations (§\ref{sec:conceptual_fewshot}), with concepts of the datasets described in §\ref{sec:datasets}. Models are ordered by a number of parameters. Error bars show a 95\% confidence interval of the bootstrapped results ($100$ samples, $200$ repeats). Absolute results for both selection strategies are in Figure~\ref{fig:perf_absolute}.
  }
  \label{fig:perf1}
\end{figure*}

\begin{figure}[th]
  \centering
    \includegraphics[width=0.43\textwidth]{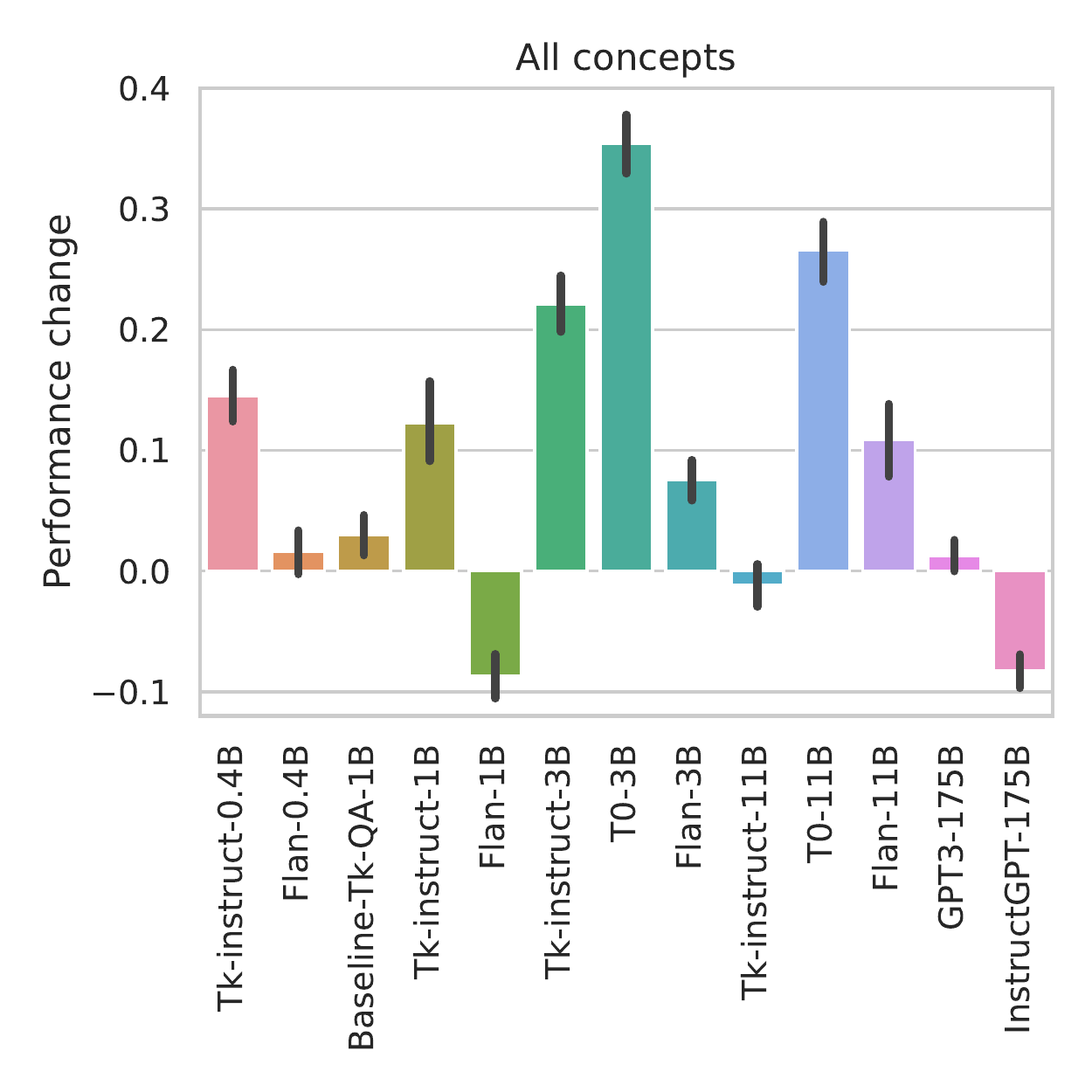}\vspace*{-3mm}\caption{\textbf{Conceptual few-shot evaluation: all concepts:} Error change of the assessed in-context learners between random demonstrations and concept-sharing demonstrations (§\ref{sec:conceptual_fewshot}) aggregated over all assessed concepts. Experimental setup is consistent with Figure~\ref{fig:perf1}. 
  }
  \label{fig:perf_avg}
\end{figure}

Following is a description of datasets that we use in Conceptual few-shot evaluation. Note that for each dataset, we highlight a single concept that we use in Conceptual few-shot evaluation as the $\mathcal{C}$ (§\ref{sec:conceptual_fewshot}).
In the case of each model and dataset, we first evaluate all templates available in Promptsource and report the gain of utilising the chosen concept for the best-performing template.

\paragraph{WorldTree} \citep{jansen-etal-2018-worldtree,xie-etal-2020-worldtree} is a collection of 5,114 science exam questions with the explanations in the form of 9,216 shared facts supporting the assignment of the correct answer.

We use the shared facts as the concepts~$\mathcal{C}$ and evaluate with the demonstrations of a maximal facts' intersection with the predicted sample.
Contrary to the other datasets, in WorldTree evaluation, we prepend the facts for all the demonstrations in the context before the demonstrations so that the model does not have to rely on the already obtained knowledge.

\paragraph{OpenBookQA} \citep{mihaylov-etal-2018-suit} is a collection of elementary-grade single-choice questions requiring common sense knowledge about the world.
A set of 4,957 explanations take the form of a triple of \textit{(object, relation, object)}, such as ``a stove generates heat'' for a question ``Which one of these can help a person cook their food? [four options]'' and a correct option ``a counter cooker appliance''.

To extract informative concepts $\mathcal{C}$, we perform syntactic analysis of the explanation and extract the \textit{relation}, identified as a \textit{root} of the sentence's parse tree.
Hence, in concept-sharing few-shot learning, we prime the aforementioned question with other question-options-answer pairs of the questions answerable by relating the input to output through the ``generate'' function.

\paragraph{HotpotQA} \citep{yang-etal-2018-hotpotqa} is a QA dataset composed of questions requiring the QA model to jointly reason over multiple passages of multi-document contexts. \citet{inoue-etal-2020-r4c} enrich the dataset with explanations from three human annotators. The explanations are structured in the form of triples $(e_1, r, e_2)$, associating two entities ($e_1$ and $e_2$) through a \textit{relation} $r$, such as (``Scott Derrickson'', ``is'', ``an American director'').

We extract the shared concepts $\mathcal{C}$ as pairs of $(r, e_2)$; Hence, Conceptual few-shot will prime the prediction with questions and contexts presenting the entities in analogical relations to the ones the model should understand for correct prediction.

\paragraph{GLUE Diagnostic} \citep{wang-etal-2018-glue} contains approximately 1,100 diagnostic samples of Natural Language Inference intended to fool a simple statistical model. While the concepts are heuristically extracted in other cases, GLUE diagnostic directly annotates 30~distinct logical concepts needed in prediction, such as \textit{double negation}, \textit{conjunction}, or \textit{existential quantification}.
We directly use these logical concepts as the reasoning concepts $\mathcal{C}$.

\subsection{Baseline model (\textsc{Baseline-Tk-QA-1B})}


To contextualize the results of existing in-context learners, we additionally evaluate a simple newly-created few-shot in-context learner trained on a single QA dataset. Similarly to \textsc{Tk-Instruct}, we construct the training examples of the meta-learning task in the explicit few-shot learning format, as initially proposed by \citet{min-etal-2022-metaicl}, where the model is updated to predict correct labels with a set of randomly-selected demonstrations included in the input (Eq.~\eqref{eq:few_shot}). This way, we fine-tune a \textsc{T5-Large} model \cite{t5} on AdversarialQA dataset \cite{bartolo-etal-2021-improving} until convergence on a validation split. We assess the resulting model on Concept-sharing few-shot learning together with other in-context learners and denote its results as \textsc{Baseline-Tk-QA-1B}.

\section{Results and Discussion}

Figure~\ref{fig:perf1} shows a relative change in models' performance between a random selection of demonstrations and Concept-sharing few-shot learning, i.e. with demonstrations sharing a selected concept (§\ref{sec:conceptual_fewshot}), ordered by models' size.

For each of the assessed concepts, we observe statistically significant improvement for at least one of the models, which confirms our initial assumption on the informativeness of the extracted concepts in prediction.

Following the ordering of the results, we see that concept learning does not relate to the model size.
Following the results for each model instance, we see many cases where the model is able to utilise one concept but fails to utilise, or even worsens the prediction when exposed to the other.
The variance is larger for instruction-tuned \textsc{Tk-instruct} models, excelling in utilising shared reasoning logic of GLUE, but to the contrary, degrading when being exposed to fact-sharing demonstrations of WorldTree. Contrary to these results is the case of \textsc{InstructGPT} that is agnostic to concepts except for GLUE. We speculate that this could be explained by the exposition of evaluation samples in training.

Figure~\ref{fig:perf_avg} shows the average of changes of Conceptual few-shot evaluation over the inspected four concept types. 
The aggregation uncovers that the gain from providing informative demonstrations largely varies among models, with \textsc{T0} and larger \textsc{Flan} models benefiting from the presented concepts more often. We analyse these two cases and find that \textsc{Flan}'s concept learning gains need to be mainly attributed to decreased performance on unseen tasks in random-demonstrations settings (Figure~\ref{fig:perf_absolute}), thus making this a disputable success. However, in the case of \textsc{T0}, we find both model versions able to benefit from the concept in all (8) cases with a single exception of \textsc{T0-11B} in OpenBookQA. Contrary to other instructional models trained on mixtures of over 1,600 tasks, \textsc{T0} was trained on a significantly smaller mix of 35 tasks of mainly QA category with open domain of both input prompts and labels. Therefore, we speculate that the superiority of \textsc{T0} in concept learning may attribute to a reliance of vast multi-task learners on spurious features such as the mapping of the evaluation task to some of the previously seen tasks.
\section{Conclusion}
This work introduces a Concept-sharing few-shot learning task that reflects on in-context learners' ability to extract a specific reasoning concept from demonstrations and apply it in a prediction. We assess a set of recent in-context learners for this ability over a set of concepts extracted from human explanations. 

While we find each of the evaluated concepts to be informative for at least one model, we show that most in-context learners can not benefit consistently from all concepts. 
Despite that, we still observe some intriguing trends, such as that the concept-learning ability does not appear to relate to the model size, or that the \textsc{T0} models are able to benefit from concepts much more consistently than the models maintaining vastly larger mixtures of pre-training tasks.

We believe that future work can inspire in the proposed approach of a more detailed evaluation of the models' behaviour. We trust that similar evaluations, possibly scaled to a more comprehensive selection of reasoning concepts, will allow us to better understand the capabilities of universal language models and to refine our expectations of their behaviour accordingly.


\section*{Limitations}

\paragraph{Concepts}

In this work, we extract the concepts from  semi-structured explanations whose format reassures consistency and non-ambiguity of the exploited concept(s).
The selection of datasets and corresponding \textit{concepts} is primarily conditioned by data availability, as the semi-structured explanations are available merely for a small set of datasets.

We acknowledge that our selection of concepts is not representative for a vast variance of concepts that users might expect models to learn from context in interaction.
Some important concepts' features that we identify are following:
\textbf{(i)} a number of premises or reasoning inference steps needed to map the input to output,
\textbf{(ii)} the granularity of the reasoning steps,
\textbf{(iii)} a type of the premises; For instance, whether the familiarity with a given concept requires a \textit{memorization} of an entity property (such as ``\textit{sun emits light}''), or a \textit{reasoning mechanics} such as analogical reasoning (``\textit{if animals can run and cat is an animal, then a cat can run}'').

We invite future work to identify or propose a taxonomy that would better reflect the wide variance of reasoning concepts that models are expected to comprehend in order to serve a wide scope of unseen tasks. Such taxonomy can motivate a more targeted collection of concepts from explanations, or annotation of new explanations demonstrating new concepts.

\paragraph{Models}
We acknowledge the limitation in a variance of evaluated models given by their availability and our computational possibilities. We evaluate only two models of the GPT family due to the usage limits of OpenAI API. Outside GPT models, we do not evaluate models over 20B parameters, given the infrastructure requirements of such settings. Nevertheless, we argue that the relevance of the models with constrained access, or resource requirements exceeding the limits of most organizations also remains a subject of open question.

\paragraph{Datasets}
One should note that the sizes of our evaluation datasets, for which we are able to extract concepts from explanations (Fig.~\ref{fig:perf1}), might be too small to compare concept sensitivity in the absolute numbers. The sizes of our sensitivity evaluation datasets are the following: WorldTree: 2,204 samples, OpenBookQA: 792 samples, GLUE Diagnostics: 282 samples, HotpotQA: 182 samples.

\section*{Ethical Considerations \& Broader Impact}



As outlined in Section~\ref{sec:intro}, in-context learning recently presents a research direction of broad public interest, where the outstanding results on NLP benchmarks often do not meet the users' expectations.
It is understandable that the focus of development in in-context learning LLMs goes to measurable improvements on existing benchmarks, as ecologically-valid evaluations on end use-cases are timely and difficult to compare to previous work \citep{Vries2020TowardsEV}.

Nevertheless, in this highly-exposed and fast-paced direction of research, we identify the necessity for the emergence of fast proxy measures that can shed light on the decision-making of the LLMs as expected by their end users. 

The presented evaluation of models' sensitivity to demonstrated reasoning concepts introduces a technical framework for quickly assessing models' compliance with our expected functioning; However, a selection of a comprehensive set of concepts, that we can agree our models should be able to learn, remains a subject of open discussion.

\section*{Acknowledgements}

We acknowledge the Centre for Biomedical Image Analysis at Masaryk University, supported by MEYS CR (LM2023050 and CZ.02.1.01/0.0/0.0/18\_046/0016045 Czech-BioImaging) for their support in obtaining evaluations presented in this paper.

\bibliography{marekov,stefanik}
\bibliographystyle{acl_natbib}

\appendix

\section{Details of Concept-aware Evaluations}
\label{appx:evaluation}

\label{sec:training}
\begin{figure*}[t]
  \centering
    \hspace*{-8.1mm}
    \includegraphics[width=0.282\textwidth]{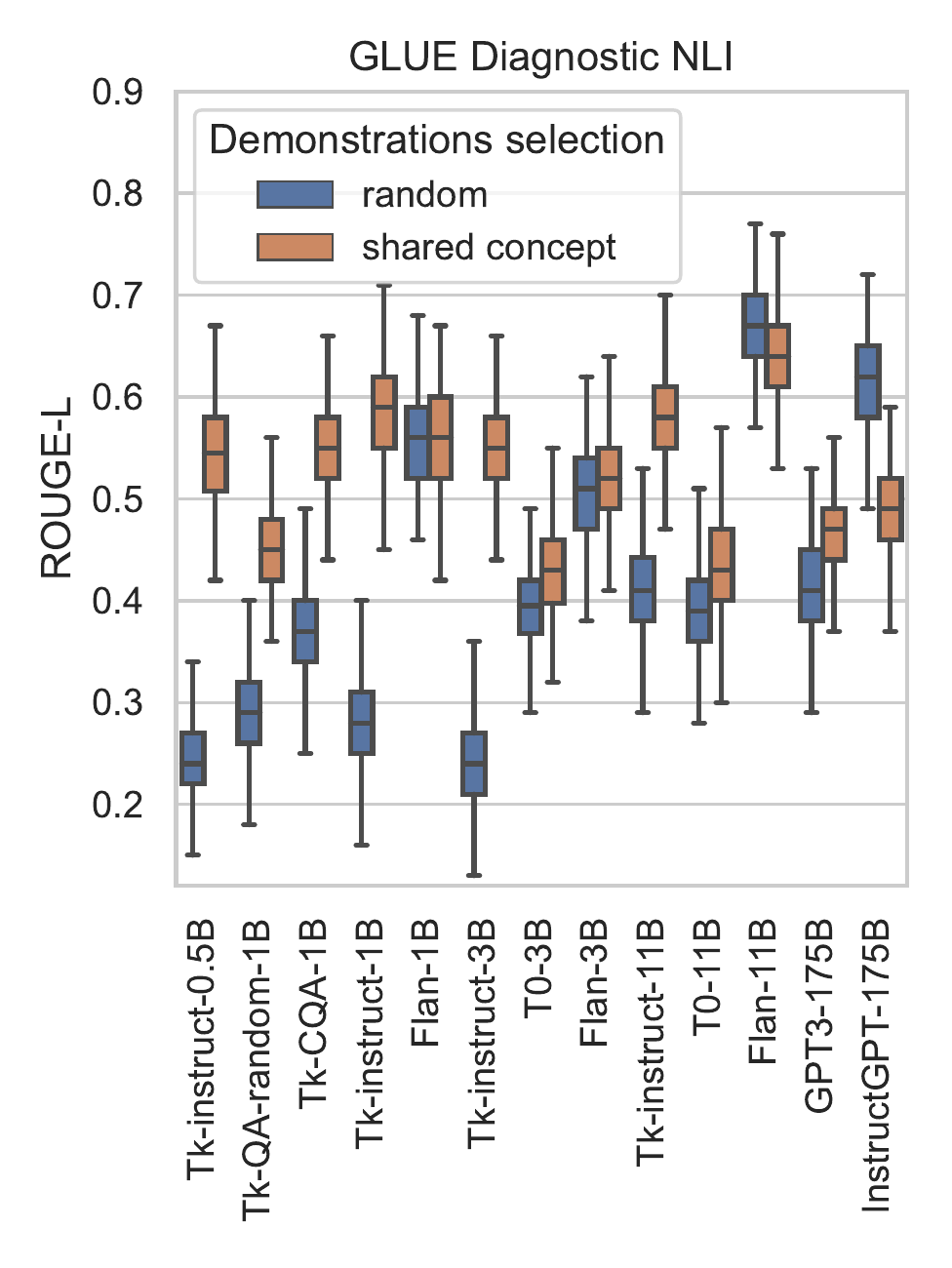} \hspace{-3mm}\includegraphics[width=0.267\textwidth]{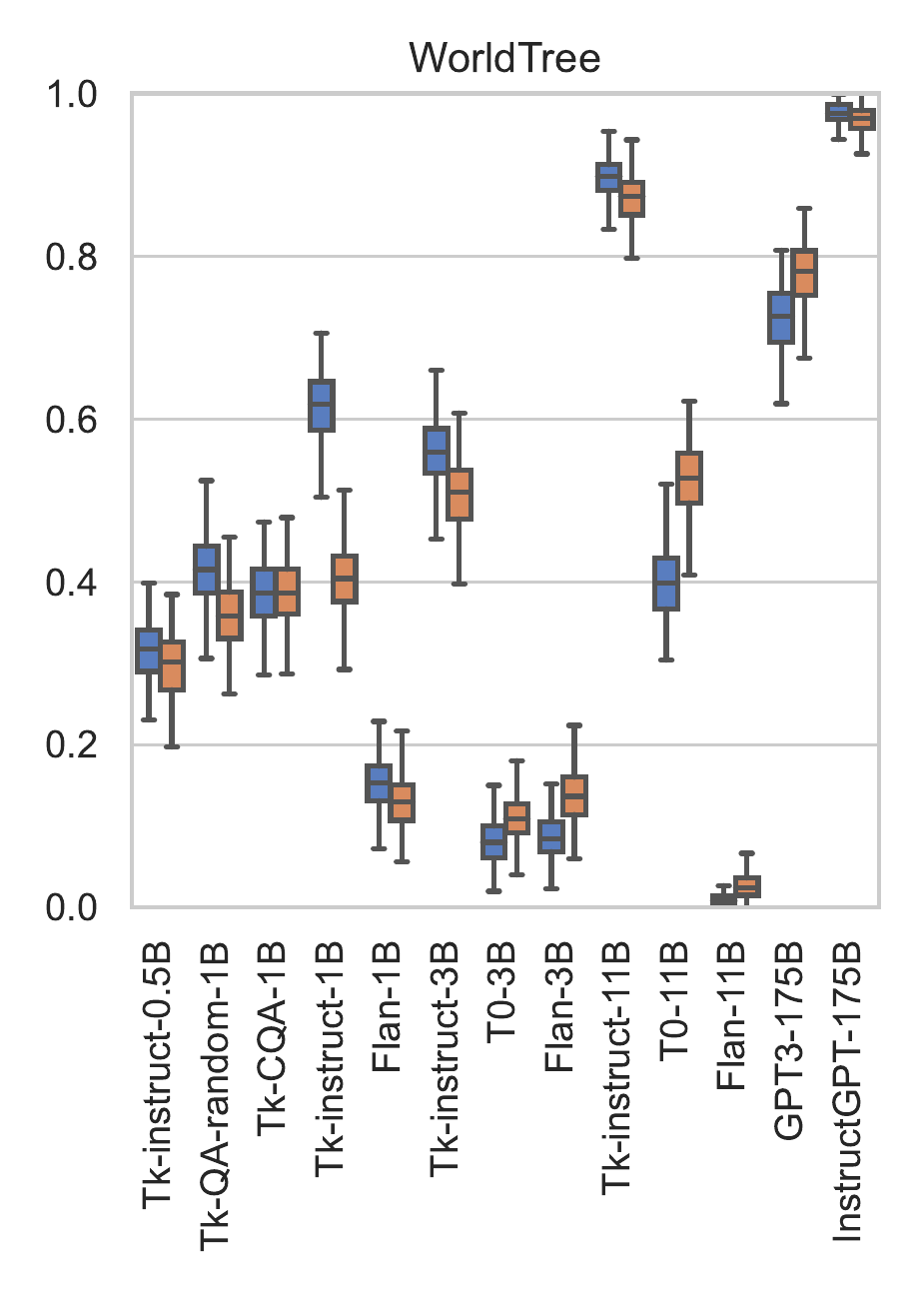} \hspace{-3mm}\includegraphics[width=0.267\textwidth]{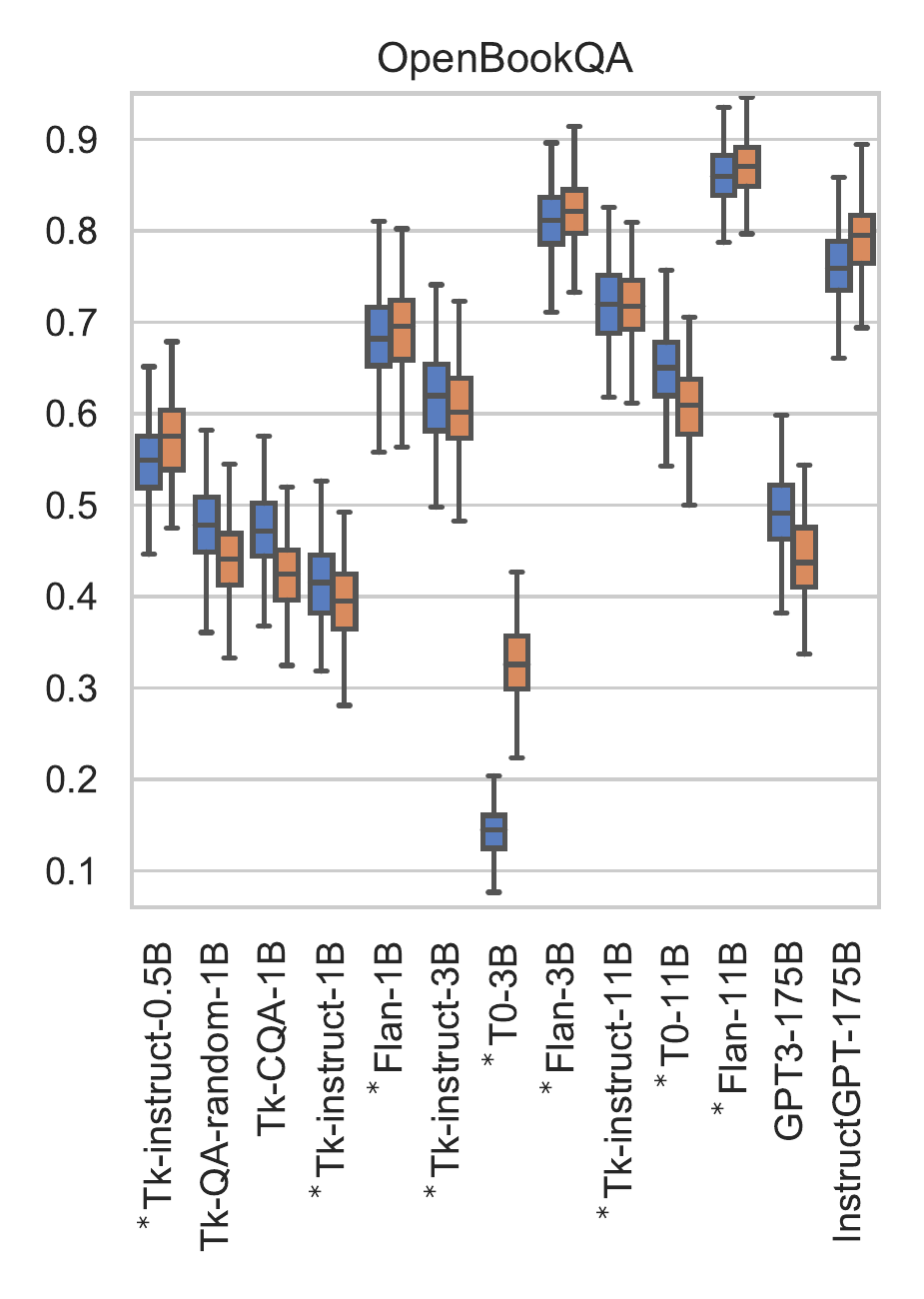} \hspace{-3mm}\includegraphics[width=0.267\textwidth]{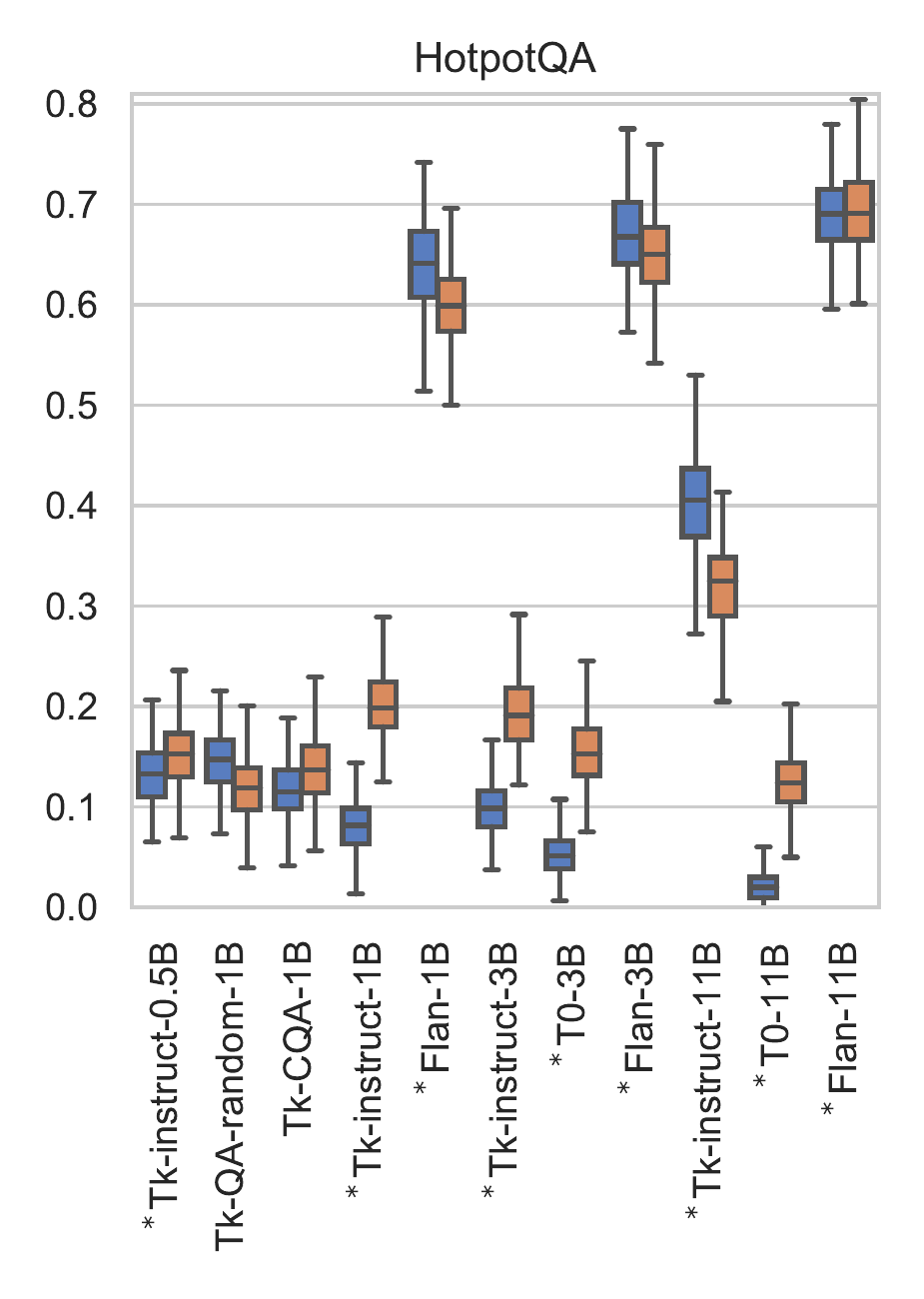} 
  
  \caption{\textbf{Conceptual few-shot evaluation:} ROUGE-L of models using \textit{random} demonstrations (left) and demonstrations exploiting a concept of prediction (§\ref{sec:datasets}; right). Boxes and confidence intervals cover 50\% and 95\% of the bootstrapped results, respectively ($100$ samples, $200$ repeats). Models marked with $^*$ were exposed to the evaluation task (but not samples) in training. Training datasets of \textsc{GPT$^*$} models are unknown.
  }
  \label{fig:perf_absolute}
\end{figure*}

Unless stated otherwise, we evaluate \textit{all} models over \textit{all} datasets and \textit{both} demonstrations selection strategies consistently for ROUGE-L in default settings of \citet{lin-2004-rouge}, using a number of demonstrations $k=3$ and contexts constructed in the following format:

\noindent \textit{``Input: $x_1$ Prediction: $Y_1$
Input: $x_2$ Prediction: $Y_2$
Input: $x_3$ Prediction: $Y_3$
Input: $x_\text{pred}$''}

Among both random and concept-sharing evaluations, we share the same $x_\text{pred}$ and only permute the demonstrations; We find cases where the filtering of predicted samples ($x_\text{pred}$) to the ones sharing a concept with sufficient amount of (3) different samples needed for demonstrations makes the task systematically easier.

We diverge from the stated configuration only in the following cases:
\begin{itemize}
    \item \textsc{Tk-Instruct-11B} and HotpotQA: we limit the evaluation contexts to at most 3.500 unique words, as we can not fit longer contexts into the memory. This might make the absolute results in this configuration overly optimistic, but still comparable within the Conceptual few-shot evaluation.
    \item \textsc{GPT} and HotpotQA: We completely exclude these evaluations given the fixed context window size of these models will exclude the $x_\text{pred}$ from prediction input in too many cases.
\end{itemize}

We choose evaluated \textsc{GPT} APIs based on OpenAI documentation\footnote{\url{https://beta.openai.com/docs/model-index-for-researchers}}, picking for \textsc{GPT} and \textsc{InstructGPT} models marked as \textsc{davinci} and \textsc{text-davinci-003}. Note that these identifiers might change in time, thus disallowing us to guarantee the reproducibility of their evaluations.

\section{Computational Requirements}

We run both training and evaluation experiments using single \textsc{NVidia A100-SXM-80GB}.
The time and computational requirements of evaluation depend largely on the size of the evaluated model; We can evaluate the models up to 11B parameters on a single \textsc{NVidia A100-SXM-80GB}. The evaluation of Concept Few-shot learning on all our datasets, together with the Random reference evaluation takes approximately 2~hours for a 1B model.

\end{document}